\pdfoutput=1
\documentclass[11pt]{article}

\usepackage[]{ACL2023}
\usepackage{times}
\usepackage{latexsym}

\usepackage[T1]{fontenc}
\usepackage[utf8]{inputenc}

\usepackage{microtype}

\usepackage{graphicx}
\usepackage{bm}
\usepackage{amsfonts}
\usepackage{svg}
\usepackage{amsmath}
\usepackage{booktabs}
\usepackage{graphicx,multirow,array}
\usepackage{mathtools}
\newcommand{\defeq}{\vcentcolon=}

\usepackage{fancyvrb}
\usepackage[inline]{enumitem}
\newcolumntype{M}[1]{>{\centering\arraybackslash}m{#1}}
\newcommand\numberthis{\addtocounter{equation}{1}\tag{\theequation}}

\title{Document-Level Supervision for Multi-Aspect Sentiment Analysis Without Fine-grained Labels}
\author{Kasturi Bhattacharjee\thanks{~~Corresponding author} \qquad
        Rashmi Gangadharaiah \\
        AWS AI Labs \\ 
        \texttt{\{kastb,rgangad\}@amazon.com}}

\begin{document}
\maketitle
\begin{abstract}
    Aspect-based sentiment analysis (ABSA) is a widely studied topic, most often trained through supervision from human annotations of opinionated texts. These fine-grained annotations include identifying aspects towards which a user expresses their sentiment, and their associated polarities (aspect-based sentiments). Such fine-grained annotations can be expensive and often infeasible to obtain in real-world settings. There is, however, an abundance of scenarios where user-generated text contains an \textit{overall} sentiment, such as a rating of 1-5 in user reviews or user-generated feedback, which may be leveraged for this task. In this paper, we propose a VAE-based topic modeling approach that performs ABSA using document-level supervision and \textit{without requiring fine-grained labels} for either aspects or sentiments. Our approach allows for the detection of multiple aspects in a document, thereby allowing for the possibility of reasoning about how sentiment expressed through multiple aspects comes together to form an observable overall document-level sentiment. We demonstrate results on two benchmark datasets from two different domains, significantly outperforming a state-of-the-art baseline.

\end{abstract}
\section{Introduction}

Given the prevalence of large volumes of opinionated user-generated text such as reviews, social media data (e.g. tweets) that are widely available today, sentiment analysis in general and aspect-based sentiment analysis (ABSA) in particular, are vital in order to gauge user preferences and sentiments. This is particularly relevant in real-world settings in which users can leave feedback on a variety of aspects about their experience in using the product, each carrying a different sentiment (Table \ref{tab:multi-aspect example}). Detecting aspects and their corresponding sentiments from such user-generated text is essential in gaining insights into user experience and identifying potential avenues of product improvement. 

ABSA is a well-studied task \cite{xue2018aspect, tang2020dependency, peng2020knowing, li2021dual, rietzler2019adapt} in opinion mining \cite{pang2008opinion, liu2012sentiment} that allows one to perform fine-grained sentiment analysis on documents. This involves detecting \textit{aspects} that occur in the text, and their associated \textit{sentiments}. For instance, in Table \ref{tab:multi-aspect example}, {\color{blue}\textbf{food}}, {\color{blue}\textbf{steak}}, {\color{blue}\textbf{fish}} and {\color{blue}\textbf{pork}} are a few aspects mentioned in a restaurant review towards which the user has expressed positive sentiment.
\begin{table}[!t]
    \small
    \begin{tabular}{M{61mm}|M{9mm}} \hline
         \textbf{Review} & \textbf{Rating} \\ \hline \hline
         Delicious healthy {\color{blue}\textbf{food}}. The {\color{blue}\textbf{steak}} is amazing. {\color{blue}\textbf{Fish}} and {\color{blue}\textbf{pork}} are awesome too. {\color{blue}\textbf{Service}} is above and beyond. Not a bad thing to say about this place. Worth every penny! & 5 \\ \hline
        
         {\color{red}\textbf{Touchpad}} didn't work, tried turning on/off but never saw a {\color{red}\textbf{cursor}} on the {\color{red}\textbf{screen}} - after several calls to customer service and doing a factory reset - which didn't work 9 out of 10 tries - CS said only 2 options, have it repaired or return it.  I'm returning it. & 1 \\ \hline

        The {\color{blue}\textbf{UI}} has been amazingly easy to use! Thanks for adding the new {\color{blue}\textbf{tabs}} on the page and enlarging the {\color{blue}\textbf{font}}! & 5 \\ \hline
        
    \end{tabular}
    \caption{Examples of reviews with multiple aspects (in \textbf{bold}) from two domains. The first is a Restaurant review, the second is a Laptop review, while the third is an example of user-generated feedback. Positive and negative sentiments associated with the corresponding aspects are denoted by the {\color{blue}blue} \& {\color{red}red} colors respectively. The Ratings column represents the overall rating or satisfaction score provided by the user.}
    \label{tab:multi-aspect example}
\end{table}

Most prior work in this area \cite{xue2018aspect, tang2020dependency, li2021dual, rietzler2019adapt, zhang2019aspect, peng2020knowing, wang2017coupled, hu2019open} approach ABSA in a supervised fashion which requires \textit{\textbf{fine-grained annotations for both subtasks, i.e. aspect and sentiment detection}}. Obtaining such annotations is time-consuming and expensive, and is often infeasible in real-world scenarios. However, there often is an overall document-level sentiment or user satisfaction score present in such texts, e.g. review ratings (Table \ref{tab:multi-aspect example}) which may be leveraged for this task.

Further, it may be posited that user sentiment expressed through multiple aspects in a document has an influence over the overall document-level rating/sentiment. For instance, in Table \ref{tab:multi-aspect example}, the user feedback with a rating of 5 contains positive sentiments towards the aspects \textbf{{\color{blue}UI}} and \textbf{{\color{blue}tabs}}. Hence, leveraging the document-level sentiment to \textit{infer} the aspects and their sentiments would be of interest. In this work, we address the real-world challenge of obtaining fine-grained annotations for ABSA by proposing a neural topic modeling approach that instead \textit{utilizes document-level sentiment supervision} to perform multi aspect based sentiment analysis. We \textbf{do not require fine-grained annotations for this task}. In particular, we propose a Variational Auto-Encoder (VAE)-based \cite{kingma2013auto} topic model that instead of relying on sentiment topics to make ABSA inferences, explicitly assigns sentiment scores to specific topics and associates aspects with a subset of these topics. Our model significantly outperforms a state-of-the-art baseline on ABSA across two domains - Restaurants and Laptops. Further, the methodology and results obtained are applicable to understanding aspects and sentiments that users express in their feedback, to curate actionable insights for our product. 

\paragraph{Contributions of our work:}
\begin{itemize}
   
    \item In order to support our real-world use case of performing ABSA \textit{without fine-grained annotations},  we propose a VAE-based topic model that leverages document-level user-reported sentiment in order to detect multiple aspects and sentiments from documents, that does not require fine-grained labels. This helps reduce annotation costs significantly, which can be crucial for industrial settings.
    \item Our approach demonstrates significant boost in performance in comparison with a state-of-the-art baseline for benchmark datasets across two domains. 
\end{itemize}

\section{Related Work}
Most prior work approach ABSA in a fully supervised fashion which do not directly apply to our problem setting. A vast majority of these approaches focus solely on fine-grained sentiment detection of %already provided 
pre-determined aspects \cite{xue2018aspect, tang2020dependency, li2021dual, rietzler2019adapt, zhang2019aspect}, while some of these perform both aspect and sentiment detection in an end-to-end fashion \cite{peng2020knowing, wang2017coupled, hu2019open}.  

A few previously proposed unsupervised approaches forgo sentiment classification and just consider aspect detection. Attention-based Aspect
Extraction (ABAE) \cite{he-etal-2017-unsupervised} uses an autoencoder to reconstruct sentences through aspect embeddings, while filtering out non-aspect words using an attention mechanism. Likewise CAt \cite{tulkens-van-cranenburgh-2020-embarrassingly} uses a POS tagger and in-domain static word embeddings for unsupervised aspect classification. \citet{Fei2021WWW} motivate the idea that latent aspect-based sentiment informs document-level sentiment. They utilize the latent encoding to guide attention over the token sequence, however their approach does not attribute specific tokens to topics. Similarly, \citet{pergola-etal-2021-disentangled} associate sentiment predictions with topics while additionally disentangling sets of opinion and plot topics. However, these individual topics are not associated with specific aspects or evaluated against labeled data.

Very few approaches explore unsupervised or weakly-supervised approaches to ABSA. Notable ones are JASen \cite{huang-etal-2020-weakly}, \citet{Lin2009JST} and \citet{Lakkaraju2011SIAM}. Work by \citet{Lakkaraju2011SIAM} uses a variant of a HMM-LDA where words have 3 hidden variables: a class (sentiment, facet, or background), a sentiment topic, or an aspect topic (depending on the class hidden variable). Again sentiment and aspect inferences are made separately. \citet{Lin2009JST} introduce a joint sentiment topic model (JST), that enables association of sentiments with latent topics, however their intention is not to interpret topics as aspects and evaluate against labeled fine-grained data. JASen \cite{huang-etal-2020-weakly} learns a joint embedding space for words, documents, and topics using small seed word sets for each topic and train on an unlabeled corpus of reviews from \citet{Zhuang2020SIGIR}. However, inference is performed only on single-aspect documents containing no neutral sentiment, which does not satisfy real-world use cases.

\section{Datasets}

\begin{table}[]
    \small
    \begin{tabular}{|M{12mm} M{13mm} M{18mm} M{18mm}|} \hline
        \textbf{Domain} & \textbf{Train} & \textbf{Dev} & \textbf{Test} \\ \hline
        Restaurants & Yelp: 100K &  SemEval: 817 & SemEval: 872 \\
        Laptops & Amazon: 80K & SemEval: 502 & SemEval: 383 \\ \hline
    \end{tabular}
    \caption{Number of samples for each data split for both domains. Training data contains only document-level ratings, \textbf{no fine-grained aspect/sentiment labels}. Details are provided in Section \ref{sec:data}.}
    \label{tab:data}
\end{table}

In this work, we utilize datasets from Restaurants and Laptops domains containing overall document-level ratings but \textbf{no fine-grained aspect or sentiment labels for training}. Labeled benchmark datasets are used solely for evaluation purposes. We are unable to share experiments and results conducted on internal datasets, owing to legal limitations, but find these benchmark datasets to be closely representative of the problem at hand.

\paragraph{With document-level ratings (for training)} We sample Restaurant reviews from the Yelp Challenge Dataset\footnote{https://www.yelp.com/dataset} and Laptop reviews from the Amazon Review Dataset \cite{ni-etal-2019-justifying} for training. Titles of the reviews are concatenated with the text for the Amazon reviews dataset. Both datasets consist of English reviews.

Reviews come with user assigned star ratings from 1 to 5. We create stratified datasets by sampling equally from the ratings. 

\paragraph{With fine-grained labels (for evaluation only)}
For our development and test sets, we utilize annotated SemEval datasets from the Restaurants and Laptops domains. Specifically, we combine SemEval 2015 Task 12 \cite{pontiki-etal-2015-semeval} and 2016 Task 5 \cite{pontiki-etal-2016-semeval} test sets to act as our held-out test sets for each domain, similar to JASen \cite{huang-etal-2020-weakly}, which we compare against. The training data from SemEval 2015 acts as our development set.

Following previous work, we only utilize the entity and not the attribute part of the aspect label to reduce sparsity. 
Unlike JASen, we do \textit{not} filter out multi aspect sentences or those with \textit{neutral} sentiment from the SemEval splits. Thus our splits contain both single and multi-aspect sentences, and all three sentiment labels, i.e. \textit{positive}, \textit{negative}, \textit{neutral}, mimicking real-world scenarios. The aspect categories for both domains are identical to those used in JASen \cite{huang-etal-2020-weakly}. Data statistics are provided in Table \ref{tab:data}.
\label{sec:data}

\section{Methodology}
\begin{figure}[!h]
\centering
\includegraphics[width=0.49\textwidth]{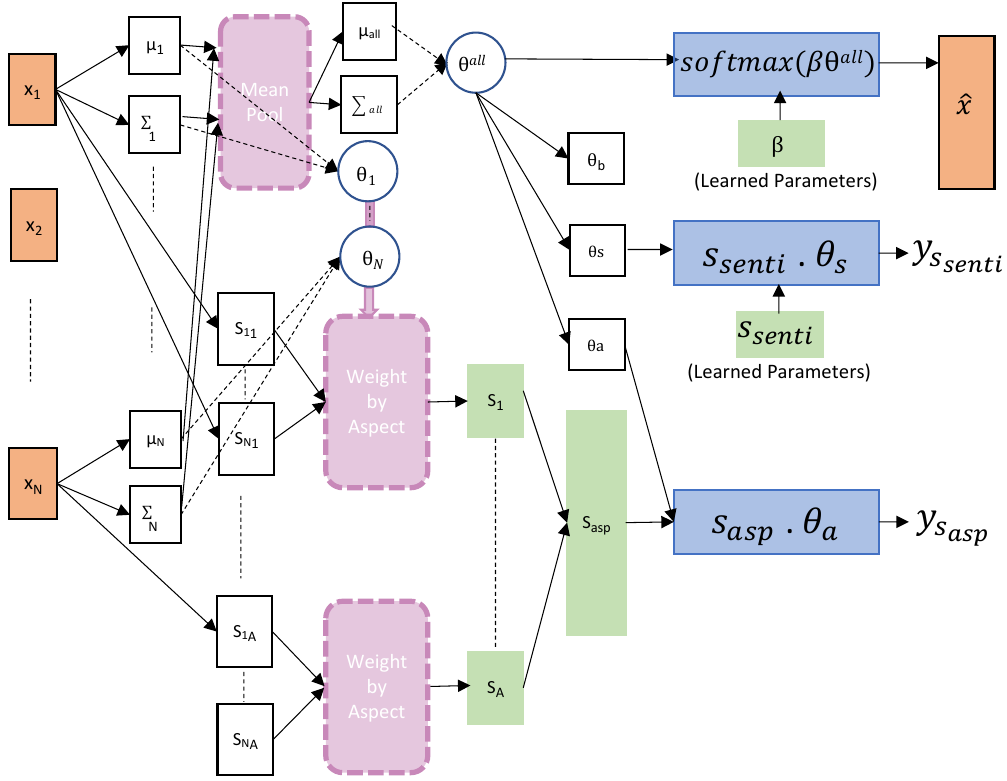}
\caption{A VAE topic model with transformer sequence inputs, BoW reconstruction and overall sentiment objectives.} 

\label{fig:architecture}
\end{figure}

The foundation of our model is a VAE \cite{kingma2013auto} based topic model following \citet{Srivastava2017ICLR} that imposes an approximation of a Dirichlet prior on the variational inference of the document-topic distribution $\bm{\theta}$. We maintain the reconstruction objective $\bm{\hat{x}}$ as a Bag-Of-Words (BoW) representation of the document to focus topics on broad semantic self-supervision. However we change the input of the autoencoder to use the sequence of token embeddings ${x^e_1} ... {x^e_N}$ produced by a pretrained transformer model (RoBERTa-base \cite{liu2019roberta} used, but could be replaced by other similar models) in order to localize the topics and sentiment scores that we discover from the document level self-supervision. We find that freezing the pretrained transformer weights is critical to performance. So we learn pooling weights, ${b} \in \mathbb{R}^L$, for the $L$ layers of a transformer model ${H} \in \mathbb{R}^{H \times L}$, where $H$ is the transformer dimensionality:
\begin{equation*}
{x^e_i} = {H_i}{b}
\end{equation*}

Beginning at the top of Figure \ref{fig:architecture}, we infer the document-topic distribution $\bm{\theta^{all}} \in \mathbb{R}^K$ where $K$ is the number of topics. $\bm{\theta^{all}}$ is the concatenation of  $\theta_a$, $\theta_s$, and $\theta_b$ which consist of $A$, $S$, and $B$ topics of aspect, sentiment, and background words respectively. 
Background topics capture words that do not apply to aspects or sentiments. The encoder produces $\bm{\theta^{all}}$ via the reparameterization trick as follows:
\begin{align*}
    {\mu_i} & = mlp({x^e_i}) \\
    {\Sigma_i} & = diag(mlp({x^e_i})) \\
    \bm{\mu_{all}} & = \frac{1}{N}\sum_{i=1}^N\bm{\mu_i} \\
    \bm{\Sigma_{all}} & = \frac{1}{N}\sum_{i=1}^N\bm{\Sigma_i} \\
    \epsilon & \sim \mathcal{N}(0,I) \\
    \bm{\theta^{all}} & = softmax(\bm{\mu_{all}} + \bm{\Sigma_{all}}^{1/2}\epsilon) \numberthis
\end{align*}

Here $diag()$ converts a vector into a diagonal matrix, and $mlp()$ is a multilayer perceptron. 

Following the ProdLDA approach of \citet{Srivastava2017ICLR}, $\bm{\theta^{all}}$ is used to produce the reconstruction as a weighted product of experts:
\begin{equation}
    %\bm{\hat{x}} = softmax(\bm{\beta}\bm{\theta})
    \bm{\hat{x}} = softmax(\bm{\beta}\bm{\theta^{all}})
\end{equation}
Here $\bm{\beta} \in \mathbb{R}^{V \times K}$ is the topic-word matrix and $V$ is the number of words in the BoW vocabulary.

Meanwhile at the bottom of Figure \ref{fig:architecture}, we take a weighted pooling ${s_{asp}}$ of token-level aspect-based sentiment judgements by also sampling token-topic distributions ${\theta_i}$ and weighting tokens-level judgements by their relative probabilities for a topic.
\begin{align*}
    \epsilon_i & \sim \mathcal{N}(0,I) \\
    {\theta_i} & = softmax({\mu_i} + {\Sigma_i}^{1/2}\epsilon_i) \\
    a_{ik} & = \frac{\theta_{ik}}{\sum_{j=1}^N \theta_{jk}} \\
    {s_i} & = mlp({x^e_i}) \numberthis \\
    s_{asp_k} & = \sum_{i=1}^N a_{ik} s_{ik} \numberthis
\end{align*}

The vector ${s_{asp}} \in \mathbb{R}^A$ is used as set of coefficients to ${\theta_a}$ which are trained to make an overall document level sentiment judgement:
\begin{equation}
    \hat{y}_{s_{asp}} = \sigma({s_{asp}} \cdot {\theta_a})
\end{equation}
Here $\sigma$ is the sigmoid function.

Meanwhile ${\theta_s}$ is used similarly to make an overall document level sentiment judgment with a set of learned parameters ${s_{senti}}$:
\begin{equation}
    \hat{y}_{s_{senti}} = \sigma({s_{senti}} \cdot {\theta_s})
\end{equation}

\subsection{Loss Functions}
We combine the same evidence lower bound (ELBO) objective used by \citet{Srivastava2017ICLR} with overall mean squared error losses for aspect and sentiment detection:

\begin{align*}
& \begin{aligned}
\mathcal{L}_{kl} & = -\frac{1}{2}\big(tr(\bm{\Sigma}_{p}^{-1} \bm{\Sigma_{all}}) \\
&+  (\bm{\mu}_{p}-\bm{\mu_{all}})^T \bm{\Sigma}_{p}^{-1}(\bm{\mu}_{p} - \bm{\mu_{all}}) \\ &- K + \log(\frac{|\bm{\Sigma}_{p}|}{|\bm{\Sigma_{all}}|})\big)
\end{aligned}\\
    & \mathcal{L}_{r} = \bm{x}^T \log(\bm{\hat{x}})\\
    & \mathcal{L}_{s_{asp}} = (\hat{y}_{s_{asp}} - y_{s})^2 \\
    & \mathcal{L}_{s_{senti}} = (\hat{y}_{s_{senti}} - y_{s})^2 \\
    & \mathcal{L} = \sum_{d=1}^{D} c_1\mathcal{L}_{kl} + c_2\mathcal{L}_{r} + c_3\mathcal{L}_{s_{asp}} + c_4\mathcal{L}_{s_{senti}} \numberthis \label{eqn:loss}
\end{align*}
\begin{table}[!t]
\small

\begin{tabular}{M{5mm}M{20mm}M{35mm}}
\textbf{Domain} & \textbf{Topic} & \textbf{Examples of Seed Words} \\ \hline\hline
\multirow{5}{*}{R} 
& aspect:\textit{food} & food pizza sushi \\

& aspect:\textit{location} & location place city\\
& aspect:\textit{service} & service waiter staff \\

& sentiment:\textit{positive} & great fresh attentive \\ 
& sentiment:\textit{negative} & rude pricey soggy \\ \cline{1-3}

\multirow{5}{*}{L} &  aspect:\textit{battery} & battery power hours \\
& aspect:\textit{display} & display screen led \\
& aspect:\textit{support} & support warranty service \\
& sentiment:\textit{positive} & easy fast lightweight \\
& sentiment:\textit{negative} & hard old slow \\ \cline{1-3}
\end{tabular}
\caption{\small{Examples of seed words used for aspect and sentiment topics across Restaurants (\textbf{R}) and Laptops (\textbf{L}) domains. Appendix Section \ref{sec:bkgrnd seeds} contains background seed words.}}
\label{tab: seed words}
\end{table}
Here $D$ is the number of documents in a batch, the variables $c$ are loss weight hyperparameters, $y_s \in [0,1]$ is a rescaled ground truth overall sentiment rank, and $\bm{x}$ is the ground truth BoW for a document. $\bm{\mu}_p$ and $\bm{\Sigma}_p$ are the mean and covariance of the multivariate normal over the softmax variables $\bm{h} = \bm{\mu_{all}} + \bm{\Sigma_{all}}^{1/2}\epsilon$ used for the approximation of the Dirichlet prior $p(\bm{\theta} | \bm{\alpha})$:
\begin{align*}
    \mu_{p_k} & = \log(\alpha_k) - \frac{1}{K} \sum_{i=1}^{K} log(\alpha_i) \\
    \Sigma_{p_k} & = \frac{1}{\alpha_k} (1 - \frac{2}{K}) + \frac{1}{K^2} \sum_{i=1}^{K}\frac{1}{\alpha_i}
\end{align*}

\subsection{Aspect and Aspect-Based Sentiment Inference}
\begin{table*}[!ht]
\centering
\small
\begin{tabular}{ccc ccc ccc}
& & & \multicolumn{3}{c}{\textbf{Aspect}} & \multicolumn{3}{c}{\textbf{Aspect-based sentiment}} \\ \cline{4-9} 
 \textbf{Domain} & \textbf{Method} &\textbf{Unlabeled Train} & P & R & F1 & P & R & F1 \\ \hline
\multirow{2}{*}{Restaurants} & JASen & \multirow{2}{*}{Yelp} & 0.50  &  0.13  &    0.21 & 0.06   &   0.01   &   0.02 \\
& \textbf{Ours} & &\textbf{0.56}  &    \textbf{0.53}  &    \textbf{0.51} & \textbf{0.15}  &    \textbf{0.20}     & \textbf{0.16} \\ \cline{1-9}

\multirow{2}{*}{Laptops} & JASen & \multirow{2}{*}{Amazon} & \textbf{0.54}  &    0.08  &    0.14 & 0.03    &  0.00   &   0.01\\ 
& \textbf{Ours} & & 0.46   &   \textbf{0.40}  &    \textbf{0.40} & \textbf{0.19}   &   \textbf{0.14}    &  \textbf{0.13} \\ \cline{1-9}
\end{tabular}
\caption{\small{Precision (P), Recall (R) and Macro-F1 (F1) for both tasks - aspect detection and aspect-based sentiment detection, on SemEval Test sets using our proposed approach, in comparison with JASen \cite{huang-etal-2020-weakly}.}}
\label{tab: quant results}
\end{table*}

In order to allow for multiple aspects per document, we make $K$ independent inferences for each aspect:
\begin{equation}
    \bm{\hat{y}_{a_{multi}}} \defeq \bm{\theta_a} > t
\end{equation}
where $t$ is a threshold hyperparameter. Finally we produce the corresponding $K$ aspect-based sentiment inferences, ignoring those corresponding to aspects not predicted:
\begin{equation}
    \bm{\hat{y}_{absa_{multi}}} \defeq \bm{s_{asp}}
\end{equation}

We use thresholds to convert the raw sentiment coefficients to sentiment labels $\{positive, neutral, negative\}$.
The aspect topics predicted can be matched to aspect labels through manual inspection of the topic-word distribution, automatic similarity measures between topic words and keywords, or through direct seeding of the topic-word distribution with keywords. We utilize the last option as it performed best in early experiments.

\subsection{Seeding Topics} \label{sec: seeding topics}

In order to encourage topics related to sentiments, aspects, and backgrounds, we utilized sets of manually curated seed words (see Table \ref{tab: seed words} for examples) to influence the initialization of the topic-word matrix $\bm{\beta}$. 

We bias the Xavier normal initialization by summing on additional weight for seed words, while allowing for $\bm{\beta}$ to be updated during training.

We consider two variants of this approach:

\paragraph{Direct Seeding} Here, we simply add a positive value $c$ to the seed words $v$ in the topics $t$ to which they are associated.
\begin{equation}
    \bm{\beta_{vt}} \defeq \bm{\beta_{vt}} + c
\end{equation}

\paragraph{Seed Bootstrapping} As seed word sets are much smaller than vocabulary size, we also experiment with effectively extending the seed word set by seeding all vocabulary words in proportion to their cosine similarity to the mean word embedding of the seed words within a trained embedding space.
We compute the mean embedding of the seed words for topic $t$ as $\bm{v_t}$ and use the embeddings of vocabulary words $\bm{u_v}$ to initialize as follows: 
\begin{equation}
    \bm{\beta_{vt}} \defeq \bm{\beta_{vt}} + c \frac{\bm{v_t}\cdot\bm{u_v}}{||\bm{v_t}||||\bm{u_v}||}
\end{equation}

\section{Experiments}
In this Section, we elaborate on the baseline we compare with, and the experimental settings used.

\subsection{Baseline}
JASen \cite{huang-etal-2020-weakly} a is state-of-the-art baseline for ABSA and has been shown to outperform other strong baselines such as ABAE \cite{he-etal-2017-unsupervised}, CAt \cite{tulkens-van-cranenburgh-2020-embarrassingly} and W2VLDA \cite{GARCIAPABLOS2018ExpertSystems}.
JASen involves learning a joint embedding space for words, documents, and topics using small seed word sets for each topic, where topic seed sets correspond to specific aspects or sentiments. A CNN is trained on the output of the predictions made using cosine similarity over the document topic embeddings and allowed to self-train on its most confident predictions. We use the same aspect categories for the two domains as in JASen, i.e. \textbf{\textit{food, ambience, location, service, drinks}} for Restaurants and \textbf{\textit{support, OS, display, battery, company, mouse, software, keyboard}} for Laptops.

\subsection{Experimental Settings \& Model Hyperparameters}
\label{sec:experiments}
Both the baseline (JASen) and our model are first trained on identical in-domain training data splits discussed in Section \ref{sec:data}.  Hyper-parameter optimization is conducted based on model performance on the dev set. The choice between the two variants of seed bootstrapping are also determined using the domain-specific dev sets. Details of the best model configurations are provided in the Appendix (see \ref{sec:Best_Restaurant_configuration} for Restaurants and \ref{sec:Best_Laptops_configuration} for Laptops).

\section{Results}
In this section, we elaborate on the results obtained across both domains using our proposed approach, while comparing with the baseline. Both quantitative and qualitative results are presented. 
\begin{table}[!h]
    \small
    \begin{tabular}{p{20mm}M{50mm}}
        \textbf{Topic} & \textbf{Model-detected Top Words per Topic} \\ \hline
        \multicolumn{2}{c}{\textbf{\textit{Domain: Restaurants}}} \\ \hline
         aspect:\textbf{food} & \textit{entree rolls calamari dish scallops}  \\
         aspect:\textbf{ambience} & \textit{atmosphere  setting ambience environment } \\
         aspect:\textbf{location} & \textit{neighborhood downtown area location} \\
         aspect:\textbf{service} & \textit{servers waitress server waiter hostess} \\
         aspect:\textbf{drinks} & \textit{drinks glass bottle beers} \\
         sentiment:\textbf{positive} & \textit{delicious great terrific wonderful}\\
         sentiment:\textbf{negative} & \textit{bland overwhelming overpriced mediocre}\\ \hline
         
         \multicolumn{2}{c}{\textbf{\textit{Domain: Laptops}}} \\ \hline
        aspect:\textbf{support} & \textit{support repair problem viruses coverage} \\
        aspect:\textbf{OS} & \textit{mac operating windows os linux} \\
        aspect:\textbf{\textit{display}} & \textit{display resolution screen colors} \\
        aspect:\textbf{battery} & \textit{usb battery charge life charger power} \\
        aspect:\textbf{company} & \textit{mac pro apple hp acer} \\
        aspect:\textbf{mouse} & \textit{key usb pad mouse touchpad} \\
        aspect:\textbf{software} & \textit{usb hard drive gb ram} \\
        aspect:\textbf{keyboard} & \textit{screen keyboard good windows battery} \\
        sentiment:\textbf{positive} & \textit{awesome perfect best easy satisfied} \\
        sentiment:\textbf{negative} & \textit{stuck horrible useless broken expensive} \\ \hline
    \end{tabular}
    \caption{\small{Model-detected top words per topic for each domain.}}
    \label{tab:restaurant topic words}
\end{table}
\subsection{Quantitative Results}

We report Precision, Recall \& Macro F1 scores in Table \ref{tab: quant results} for aspect detection \& aspect-sentiment identification, for both domains. We find our approach to significantly outperform JASen on both tasks across domains. We obtain an improvement of 30 F1 points for aspect detection and 14 F1 points for aspect-based sentiment detection in the Restaurants domain. In case of the Laptops domain, our approach shows an improvement of 26 and 12 F1 points for aspect and aspect-based sentiment detection tasks, respectively.

\subsection{Qualitative Results}
\begin{table}[]
    \small
    \begin{tabular}{p{40mm}M{30mm}} 
         
         \toprule
         \textbf{Text} & \textbf{Ground Truth (GT) \& Predictions (P)} \\ \midrule
         
         \multicolumn{2}{c}{{\centering\arraybackslash}\textbf{Domain: Restaurants}} \\ \midrule
         
         \textit{The wine list was extensive - though the staff did not seem knowledgeable about wine pairings} & \textbf{GT:} {\color{teal}(drinks:positive)}, {\color{teal}(service:negative)} \textbf{P:} {\color{teal}(drinks:positive)}, {\color{teal}(service:negative)} \\ \midrule
         
         \textit{Watch the talented belly dancers as you enjoy delicious baba ganoush that's more lemony than smoky.} & 
         \textbf{GT:} {\color{teal}(food:positive)}, {\color{teal}(ambience:positive)}
         \textbf{P:} {\color{teal}(food:positive)}, {\color{teal}(ambience:positive)} \\ \midrule
       
        \multicolumn{2}{c}{{\centering\arraybackslash}\textbf{Domain: Laptops}} \\ \midrule
        \textit{I really like Mac/Apple products because of their operating system.} & 
        \textbf{GT:} {\color{teal}(OS:positive)}, {\color{teal}(company:positive)}
         \textbf{P:} {\color{teal}(OS:positive)}, {\color{teal}(company:positive)} \\ \midrule
         
         \textit{I appreciate the relatively large keyboard (with number pad) and screen size.} & \textbf{GT:} {\color{teal}(display:positive)}, {\color{violet}(keyboard:positive)}
         \textbf{P:} {\color{teal}(display:positive)}, {\color{violet}(mouse:positive)} \\
         
        \bottomrule
    \end{tabular}
    
    \caption{\small{Qualitative examples from model predictions. Correct predictions are in {\color{teal}teal}. \textit{Mismatches} are in {\color{violet} purple}.}}
\label{tab:qualitative_examples}
\end{table}

Table \ref{tab:restaurant topic words} shows the top words detected by our model for each aspect and sentiment topic, for both domains. As is observed, our model is able to correctly identify words such as \textbf{\textit{entree}}, \textbf{\textit{scallops}} and \textbf{\textit{rolls}} as aspect \textit{food} for the Restaurants domain, while words like \textbf{\textit{display}}, \textbf{\textit{resolution}} and \textbf{\textit{colors}} are identified as aspect \textit{display} for the Laptops domain. Further, the model successfully categorizes \textit{domain-specific} sentiment words such as \textbf{\textit{bland}} as negative and \textbf{\textit{delicious}} as positive for Restaurants. Similarly, words such as \textbf{\textit{stuck}} and \textbf{\textit{horrible}} are successfully identified as negative sentiment words for the Laptops domain.

Table \ref{tab:qualitative_examples} presents multi-aspect sentence predictions from each of the best models in the two domains. For the Laptops domain, although the model yields good performance for other aspects, it tends to confuse between the \textit{keyboard} and \textit{mouse} aspects, which could be attributed to the semantic similarity between these aspect categories.

\section{Conclusion \& Future Work}
In this work, we explore the task of multi-aspect based sentiment analysis using document-level sentiment supervision. This allows us to detect multiple aspect categories per document and their corresponding sentiment, without requiring fine-grained labels for supervision. This is of great importance in real-world settings where obtaining fine-grained labels for the task is expensive. We train our model on unlabeled data from two domains and evaluate on annotated test sets from the same domains. Our neural topic model based approach significantly outperforms the state-of-the-art baseline for ABSA. Further inspection of qualitative results illustrates that our method detects meaningful topic words across both domains. Our methodology is applicable for gaining insights from user-generated feedback on our product, to motivate future product directions. Future work would involve exploring methods to improve our aspect-based sentiment detection performance. A potential direction is to incorporate a few labeled samples for the aspect-based sentiment detection task to aid the model.

\bibliography{custom}
\bibliographystyle{acl_natbib}

\appendix
\section{Appendix}
\subsection{Seed Words}\label{sec:Seed_Words}
We present here details on seed bootstrapping (Section \ref{sec: seeding topics}) and examples of background seed words. 

\subsubsection{Seed Bootstrapping Details}
Specifically we use the default settings on the Gensim implementation \cite{rehurek2011gensim} to train Word2Vec \cite{Mikolov2013} embeddings on all 4,724,471 Yelp Restaurants reviews and all 9,131,126 Computers \& Accessories Amazon reviews respectively to domain. We employ the same text preprocessing as for our BoW.

\subsubsection{Background Seed Words}\label{sec:bkgrnd seeds}
The following were seed words used for the background topics:
\begin{itemize}
    \item topic 1: fully somehow apparently since already
    \item topic 2: whatever another neither everyone someone
    \item topic 3: besides despite whether till
    \item topic 4: quite another every
    \item topic 5: might may must could
    \item topic 6: near among along across without
    \item topic 7: ok okay oh wow
    \item topic 8: yet plus either
    \item topic 9: five six ten four three two one
\end{itemize}

\subsection{Model Hyper Parameters}
\subsubsection{Configuration Variations}
\label{sec:Default_Configuration}
Here are the model hyper parameter variations experimented with. The best parameters chosen were based on the development set from both domains. 
BoW vocabulary values varied in \{2000, 20000\}. Num of epochs was varied in \{10, 20,30, 50\}. Number of aspect topics varied in \{5, 8, 10, 12, 15, 16, 30, 50, 100, \}. Activation varied between \textit{softplus} and \textit{relu}. Learning rate values varied in \{5e-3, 1e-3, 5e-4, 1e-4,1e-5\}. Loss weights (Equation \ref{eqn:loss}) were varied according to the following: \{$c_1$: 0.1, $c_2$: 0.1, $c_3$: 10.0, $c_4$: 10.0\}, \{$c_1$: 0.1, $c_2$: 0.1, $c_3$: 100.0, $c_4$: 100.0\}, \{$c_1$: 0.1, $c_2$: 0.1, $c_3$: 10.0, $c_4$: 100.0\},\{$c_1$: 0.1,  $c_2$: 0.1, $c_3$: 100.0, $c_4$: 10.0\}.
The thresholds used on the aspect sentiment coefficients when predicting 3 class sentiments was varied in \{1/8, 6/16, 1/5, 7/02/, 1/4, 1/2\}.
We experimented both with and without normalization of the ground truth BOW to distributions over words. Document-topic Dirichlet hyperparameter $\alpha$ was varied in \{1.0, 0.1, 0.07, 0.05, 0.02, 0.01\}. Seed $\beta$ values were varied in \{1, 10, 20\}.

\subsubsection{Best Restaurant configuration}
\label{sec:Best_Restaurant_configuration}

Here are the hyper-parameters that yielded the best performance for Restaurants domain.
Size of the BoW vocabulary used was 2000. Activation was softplus, optimizer used was adam (with 0.99 beta), learning rate 1e-5, number of epochs was 50 with a batch size of 16. Loss weights (Equation \ref{eqn:loss}) $c_1=0.1$, $c_2=0.1$, $c_3=10.0$, $c_4=10.0$. Number of epochs of zero LR before warmup for staggered start was 1, and number of epochs over which lr warmup occurs was 1. Number of aspect topics was 5. The best-performing threshold on the aspect sentiment coefficients when predicting 3 class sentiment was 1/5.  Document-topic Dirichlet hyperparameter $\alpha=1$. Seed bootstrapping gave better performance over direct seeding (Section \label{sec: seeding topics}). Not normalizing the ground truth BOW to distributions over words worked better. Seed $\beta$ value of 10.0 was used.

\subsubsection{Best Laptops configuration}
\label{sec:Best_Laptops_configuration}

Here are the hyper-parameters that yielded the best performance for the Laptops domain.
Size of the BoW vocabulary used was 2000. Activation was softplus, optimizer used was adam (with 0.99 beta), learning rate 5e-4, number of epochs was 30 with a batch size of 16. Loss weights (Equation \ref{eqn:loss}) $c_1=0.1$, $c_2=0.1$, $c_3=10.0$, $c_4=10.0$. Number of epochs of zero LR before warmup for staggered start was 1, and number of epochs over which lr warmup occurs was 1. Number of aspect topics was 8. The best-performing threshold on the aspect sentiment coefficients when predicting 3 class sentiment was 3/16.  Document-topic Dirichlet hyperparameter $\alpha=1$. Direct seeding performed better than seed bootstrapping (Section \label{sec: seeding topics}). Not normalizing the ground truth BOW to distributions over words worked better. Seed $\beta$ value of 10.0 was used.

\end{document}